\documentclass[
groupName=AIRO,      
bottomLogo=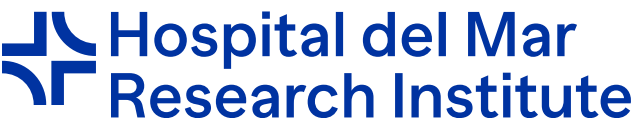,
bottomLogoH=30pt,    
]{idlab}             
\usepackage{balance} 
\usepackage[square,numbers]{natbib}
\usepackage{amsmath}
\usepackage{booktabs}
\newcommand{\aspic}{ASPIC$^{+}$}
\title{A Contractualist Argumentation Framework for Moral Decision-Making}
\shortAuthor{Marcos-Vidal et al.} 
\author[1]{Luis Marcos-Vidal}
\author[2]{Giulio Antonio Abbo}
\author[2]{Tony Belpaeme}
\affil[1]{\textit{Hospital del Mar Research Institute}, Barcelona, Spain}
\affil[2]{\textit{IDLab-AIRO}, \textit{Ghent University -- imec}, Ghent, Belgium}
\affil[ ]{\texttt{lmarcos1@researchmar.net} \texttt{giulioantonio.abbo@ugent.be}}
\date{August 2026}
\venue[To appear in the proceedings of the]{Fourth International Workshop on Value Engineering in AI (VALE 2026)}
\begin{document}

\maketitle

\begin{abstract}
Autonomous agents operating in shared environments must make decisions that affect multiple individuals with potentially conflicting interests. We propose a formal framework for moral decision-making grounded in Scanlon's contractualism, an ethical theory that evaluates the permissibility of actions in terms of principles that no one could reasonably reject. To operationalise contractualist reasoning, we use \aspic{}, a structured argumentation framework, extended with value-based filtering to model how each agent's values determine which reasons are morally relevant in the first place. The result is a Contractualist Argumentation Framework in which agents' reasons are formally represented, compared, and evaluated through argumentation semantics. We illustrate the approach through a worked example in a domestic setting and discuss its relation to existing value-based argumentation approaches.
\end{abstract}

\section{Introduction}

In shared living environments, humans constantly negotiate competing interests and justify their actions to one another.
Autonomous agents -- smart appliances today and social robots tomorrow -- operating in such settings face the same demand, yet lack principled tools to meet it.
The ethical and interpersonal concerns this raises are well documented~\cite{abbo2025concernsvalues}: when a home assistant must decide whether to disclose sensitive information, intervene in a conflict, or prioritise one person's needs over another's, the legitimacy of its decision depends not only on the outcome but on whether it can be justified to all parties involved.
Prior work has shown that agents powered by large language models can reason about social context and human values from perceptual input~\cite{abbo2026multimodal-large-language}, yet what remains missing is a principled normative framework that structures this reasoning formally.

We propose to ground such reasoning in Scanlon's contractualism~\cite{Scanlon1998}, an ethical theory that evaluates the permissibility of actions in terms of principles that none of the affected parties could reasonably reject.
This framework is particularly suited to home environments, as it is explicitly interpersonal, respects individual autonomy, and provides a procedure for comparing the reasons of different agents.
To operationalise it, we use \aspic{}~\cite{modgil2014aspic}, a structured argumentation framework in which reasons can be formally represented, combined, and evaluated.
We further draw on Value-based Reasoning Frameworks (VBFs)~\cite{wallner2024value} to model how each agent's values determine which reasons are relevant in the first place.
The result is a formal account of contractualist moral reasoning that is transparent, agent-relative, and computable.

The paper is structured as follows.
Section~\ref{sec:contractualism} introduces the contractualist framework.
Section~\ref{sec:aspic} recalls the necessary notions of \aspic{} and value-based reasoning.
Section~\ref{sec:implementation} describes how contractualist reasoning is modelled in \aspic{}.
Section~\ref{sec:example} illustrates the approach through a worked example.
Section~\ref{sec:discussion} situates our contribution with respect to related work and outlines its limitations.

\section{The Contractualist Approach}\label{sec:contractualism}

Contractualism is an ethical theory that assesses the permissibility of actions in terms of principles that could be agreed upon by free and equal individuals~\cite{RAWLS1971,Scanlon1998}. In this article, we adopt Thomas Scanlon’s version of contractualism, which explicitly limits its scope to the interpersonal domain -- what is often called morality in the narrow sense \cite{Mackie1977}. On Scanlon’s account, an action is permissible if it can be justified to others on the basis of principles that no one could reasonably reject \cite{Scanlon1998,Scanlon2008}. Moral motivation thus stems from the demand to justify one’s conduct to others, taking their standpoints and interests seriously alongside one’s own. This framework places significant weight on individual autonomy by recognising a wide range of subjective interests, including those that extend beyond personal well-being, such as concern for others. These features make Scanlonian contractualism particularly well-suited for social applications -- such as social robots or chatbots -- because it provides a structured way to model and evaluate interpersonal justification.

A central mechanism in Scanlon’s contractualism is the idea of reasonable rejection, which determines whether a principle can serve as a valid basis for moral agreement. The reasons agents may invoke in rejecting principles are subject to three key constraints: they must be \emph{personal}, \emph{individual}, and \emph{generic} \cite{Scanlon1998}. First, they are personal: because contractualism is a theory of interpersonal morality, it excludes impersonal considerations that do not relate to the interests of individuals affected by the situation at hand. Second, they are individual: objections must be grounded in the impact of a principle on particular persons, rather than in aggregative considerations such as overall welfare. Third, they are generic: reasons must appeal not to the specific circumstances or idiosyncratic preferences of a given individual, but to the standpoint of anyone who might occupy a relevantly similar position -- such as harm or a violation of privacy --  since the aim is to identify principles suitable for general regulation. Accordingly, a valid reason is one advanced by an individual on their own behalf, but framed in terms of how any person in that position would be affected by the principle under consideration. Together, these constraints give individual objections the structure they need to count as legitimate reasons.

A contractualist evaluation of an action requires a comparison of the various reasons held by different individuals. Multiple authors have suggested that the comparison of individual reasons can be understood as a process of negotiation \cite{Levine2024}. When explicit negotiation is not feasible, agents instead simulate a form of hypothetical agreement by considering what all affected parties could accept -- engaging in what is known as Virtual Bargaining \cite{Misyak2014b,Misyak2014,Chater2022,Levine2024}. This process closely mirrors the contractualist test of reasonable rejection, as it requires agents to assess actions from each individual standpoint and determine whether any party would have sufficient reason to reject the relevant principle. Virtual bargaining thus provides a tractable bridge between the normative structure of contractualism and its practical implementation, grounding interpersonal justification in the comparison of reasons across individuals. And to be valid, these reasons must satisfy the contractualist constraints of being personal, individual, and generic. To that aim, structured argumentation frameworks offer a natural formal counterpart: they can model virtual bargaining as the exchange and evaluation of competing reasons that conform to these constraints, thereby operationalising contractualist reasoning.

\section{Notions of \aspic{}}\label{sec:aspic}

\aspic{}~\cite{modgil2014aspic} is a structured argumentation framework that builds arguments from an initial knowledge base following a set of rules, describes how contrasting arguments are associated with each other, and defines how a set of arguments can \emph{win} over the rest.
We informally recall its core notions.

\paragraph{Language and Knowledge Base.}
\aspic{} operates over a formal language $\mathcal{L}$, a set of literals.
We can define a \textit{contrary} function $\overline{\phantom{x}} : \mathcal{L} \to \mathcal{L}$, whose definition could for instance be based on the negation, $\neg$.
A knowledge base $\mathcal{K} \subseteq \mathcal{L}$ represents the initial information available for argument construction.
$\mathcal{K}$ is partitioned into two disjoint subsets: $\mathcal{K}_n$, undisputable axioms, and $\mathcal{K}_p$, ordinary premises; these latter may be challenged through argumentation.

\paragraph{Rules.}
A rule $a_1, \ldots, a_n \Rightarrow b$ represents an inference from antecedents (\texttt{body}) $a_1, \ldots, a_n \in \mathcal{L}$ to a consequence (\texttt{head}) $b \in \mathcal{L}$.
The set of all rules is denoted $\mathcal{R}$ and, similarly to $\mathcal{K}$, is partitioned into strict rules and defeasible rules ($\mathcal{R}_d$), depending on whether the inference they represent is necessarily valid or potentially defeasible through argumentation.
Each rule is associated with a name via a naming function $n: \mathcal{R}_d \to \mathcal{L}$, implying that a rule can have as conclusion the negation of another rule's name, which is the basis for undercutting attacks, presented below.

\paragraph{Arguments.}
We call \emph{argumentation theory} (AT) the tuple $T = (\mathcal{L}, \mathcal{R}, n, \overline{\phantom{x}}, \mathcal{K})$.
From this, we can informally define the set of all arguments $\mathcal{A}$ inductively as follows.
A single element $x \in \mathcal{K}$ constitutes a trivial argument with conclusion $x$; further arguments are built by chaining rules whose premises are conclusions of prior arguments. We refer to the set of premises of an argument $A$ as $\texttt{Prem}(A)$ and its conclusion as $\texttt{Conc}(A)$.

\paragraph{Attacks.}
In an AT, arguments can be in conflict with each other, and this is captured by the notion of \emph{attack}, which can be of three types.
\emph{Undermining} occurs when an argument's conclusion is the negation of an ordinary premise of another.
\emph{Rebutting} occurs when an argument's conclusion contradicts the conclusion of another, and this uses a defeasible rule.
\emph{Undercutting} targets the name of a defeasible rule used in a sub-argument, defeating the rule application itself rather than its conclusion.

\paragraph{Preferences and Defeats.}
\aspic{} allows for the incorporation of preferences over arguments, which can be used to resolve conflicts when multiple arguments attack each other.
Given an ordering over arguments $\prec$, we define a \emph{successful undermining} as an attack where the attacking argument is not less preferred ($\not \prec$) than the attacked premise.
A \emph{successful rebutting} is an attack where the attacking argument is not less preferred than the argument whose conclusion is attacked.
A defeat is defined as either an undercut, or a successful undermining, or a successful rebutting.
We denote the defeat relation with $\mathcal{D} \subseteq \mathcal{A} \times \mathcal{A}$.
The tuple $F = (\mathcal{A}, \mathcal{D})$ is called an \emph{argumentation framework} \cite{dung1995acceptability}.

\paragraph{Defenses.}
A set of arguments defends an argument $A$ if for every argument $B$ that defeats $A$, there is an argument in the set that defeats $B$.
A set of arguments is \emph{conflict-free} if there is no argument in the set that defeats another argument in the set, and \emph{admissible} if it also defends all its members against outside attacks.
An admissible set of arguments is \emph{complete} if it contains all the arguments it defends.
We focus on the \emph{grounded extension} of $F$, the unique set-inclusion-minimal complete subset of $\mathcal{A}$.

\section{Notions of VBFs}\label{sec:vbfs}

So far, the AT $T$ and its associated framework $F$ are built from a shared knowledge base $\mathcal{K}$, treating all literals and rules equally.
However, especially in multi-agent settings, agents may have different perspectives on which premises and rules are relevant or acceptable, often influenced by their values.
VBFs~\cite{zurek2022towards} can be integrated in \aspic{} to address this issue at the level of $\mathcal{K}$ itself, as formalised in \cite{wallner2024value}.
Assume a set of values \texttt{Values}, a set of propositions \texttt{Prop}, and a set of agents \texttt{Agents}.
The propositions represent facts that may be relevant with respect to the values.
We also define a totally ordered set \texttt{Scale} and use it to define $\texttt{Weight} = \texttt{Scale} \cup \{?\}$.
The set \texttt{Weight} is then used to represent different degrees of value promotion, violation, or indifference (represented by ``$?$'') associated with propositions in \texttt{Prop}, as well as the value preferences of the different agents, as reported below.
A VBF is defined as a tuple $V=(\texttt{Agents},\texttt{Prop},\texttt{Values},\texttt{Scale})$

\paragraph{Value Preferences.}
VBFs define a function $\texttt{ValLimit}: (\texttt{Agents} \times \texttt{Values}) \rightarrow \texttt{Weight}$, which associates each agent and value with a tolerance level for that value.
Intuitively, the higher this threshold, the greater the degree to which a proposition must promote or violate the value in order to be considered relevant by the agent.
Similarly, a function $\texttt{ValProp}: (\texttt{Agents} \times \texttt{Values} \times \texttt{Prop}) \rightarrow \texttt{Weight}$ captures the extent to which an agent perceives a proposition to be promoting or violating a value.
Importantly, these assessments may vary across all dimensions: different agents may perceive the impact of the same proposition differently, and a single agent may evaluate the same proposition differently depending on the value under consideration.

\paragraph{Filtering.}
A literal passes the filter for an agent $\alpha \in \texttt{Agents}$ if its weight meets the limit for every value: $\texttt{PropBaseClean}_\alpha = \{p \in \texttt{Prop} \mid \forall v \in \texttt{Values}: \texttt{ValProp}(\alpha,v,p) \not < \texttt{ValLimit}(\alpha, v)\}$, with ``$?$'' causing $p$ to pass the filter.
The intuition behind this is to exclude premises that $\alpha$ would not be willing to consider in a discussion; consequently, being deemed inappropriate with respect to a single value is sufficient to discard the premise.

\paragraph{VBFs in \aspic{}.}
The set of literals passing all filters determines a \emph{subjective} knowledge base $\mathcal{K}_\alpha \subseteq \mathcal{K}$ which does not contain all discarded ordinary premises, and instead contains their negation, as defined below (from ~\cite{wallner2024value}).

\begin{equation}
\label{eq:wallner}
\begin{split}
\overline{\texttt{PropBaseClean}}_\alpha = \texttt{Prop} \setminus \texttt{PropBaseClean}_\alpha \\
\texttt{CompsProps}_\alpha = \{\neg p \mid p \in \overline{\texttt{PropBaseClean}}_\alpha\} \\
\mathcal{K}^\alpha_p = (\mathcal{K}_p \setminus \{\overline{\neg p} \mid \neg p \in \texttt{CompsProps}_\alpha\}) \\\cup \{\neg p \mid \neg p \in \texttt{CompsProps}_\alpha\} \\
\mathcal{K}_\alpha = \mathcal{K}_n \cup \mathcal{K}^\alpha_p
\end{split}
\end{equation}

Accordingly a subjective set of defeasible rules $\mathcal{R}^\alpha_d \subseteq \mathcal{R}_d$, retains only those rules whose premises and conclusion lie in $\texttt{PropBaseClean}_\alpha$.
Together these concepts define a \emph{subjective argumentation theory} and \emph{framework}.
Attacks can then arise across agents' subjective theories, and collectively acceptable arguments are defined as the intersection of the grounded extensions of each agent's subjective framework.

\section{Contractualism in \aspic{}}\label{sec:implementation}

In this section, we describe how to use \aspic{} for moral evaluation based on contractualism and virtual bargaining. We propose a framework that reasons exclusively about moral information, such that the only contestable elements in the corresponding AT are moral in nature -- for example, whether an agent objects to being harmed because it violates its wellbeing. In this way, we separate reasoning about what is morally right or wrong from reasoning about whether the underlying facts are true.

The central idea is to model a hypothetical negotiation among the agents affected by a given action. To this end, we generate arguments representing the different reasons each agent may have for or against the action. Once these arguments have been constructed, the decision regarding the action depends on whether one agent’s reasons outweigh those of another agent. Importantly, the number of reasons is irrelevant, since, from a contractualist perspective, reasons cannot be aggregated.

To describe the complete process, we divide it into three steps. First, we address the derivation of the moral reasons used to construct the arguments. These reasons must support arguments that are legitimate from a contractualist standpoint. Second, we discuss the specific requirements that the AT must satisfy in order to incorporate contractualist arguments grounded in those reasons. Finally, we examine how the different arguments are compared in order to reach a conclusion.

\subsection{Generating Moral Reasons}

To be considered legitimate from a contractualist perspective, the reasons underlying arguments must satisfy three conditions: they must be individual, generic, and personal. That means they must represent the interests of a specific agent, be grounded in principles that anyone in a similar situation could appeal to, and genuinely reflect the perspective of the individual concerned. As principles every one could appeal to, we use values because their higher level of abstraction allows a small set of values to capture a wide range of situations. Consequently, each reason needs to represent whether an action promotes or violates a value for one agent, and from the subjective perspective of that agent. 

We borrow \texttt{ValProp} and \texttt{ValLimit} from \cite{wallner2024value}; however, since in our framework we aim to identify the moral reasons that each agent may use to construct arguments for or against an action, each value--proposition pair must be evaluated independently.
This allows us to determine whether a proposition is morally relevant with respect to a particular value and should therefore be included in the argumentation process. Accordingly, we define the set of possible moral reasons associated with each proposition, value, and agent as $\texttt{MoralProp} = \{p_v(\alpha) \mid p \in \texttt{Prop},\ v \in \texttt{Values},\ \alpha \in \texttt{Agents}\}$. We then define a single $\texttt{PropBaseClean}_*$ for all agents as follows.

\begin{equation}
\begin{split}
\texttt{PropBaseClean}_* = \{p_v(\alpha) \in \texttt{MoralProp} \mid\\
\texttt{ValProp}(\alpha,v,p) \not < \texttt{ValLimit}(\alpha, v)\}
\end{split}
\end{equation}

Our $\texttt{PropBaseClean}_*$ captures which values are promoted or violated by proposition $p$ to a sufficient degree for inclusion in the argumentation process by an agent.
Note that a single proposition may give rise to multiple moral reasons if it promotes or violates several values strongly enough. This enables each $p_v(\alpha)$ to be used as an independent premise in \aspic{}, thereby allowing different argument structures to be defined for different values. This filtering approach is consistent with Scanlon’s contractualism, since its goal is to generate all possible objections that agents could raise and then determine whether any objection is sufficiently strong, relative to the others, that it could not reasonably be rejected.

The set of ordinary premises $\mathcal{K}^*_p$ is constructed as $\mathcal{K}^\alpha_p$ in Equation~\ref{eq:wallner}.

\begin{equation}
\begin{split}
\overline{\texttt{PropBaseClean}}_* = \texttt{MoralProp} \setminus \texttt{PropBaseClean}_* \\
\texttt{CompsProps}_* = \{\neg p_v(\alpha) \mid p_v(\alpha) \in \overline{\texttt{PropBaseClean}}_*\} \\
\mathcal{K}^*_p = (\mathcal{K}_p \setminus \{\overline{\neg p_v(\alpha)} \mid \neg p_v(\alpha) \in \texttt{CompsProps}_*\}) \\\cup \{\neg p_v(\alpha) \mid \neg p_v(\alpha) \in \texttt{CompsProps}_*\} \\
\mathcal{K}_* = \mathcal{K}_n \cup \mathcal{K}^*_p
\end{split}
\end{equation}

Rule filtering can be performed as in \cite{wallner2024value}, retaining only those rules whose premises and conclusion lie in $\texttt{PropBaseClean}_*$.
Differently from their approach, we do not build a subjective AT for each agent, but a single one containing all the subjective propositions. As explained below, each of these propositions will generate a subjective argument built on the single AT.
Note that we foresee the existence of inference rule schemes~\cite{modgil2014aspic} $r(x): \varphi(x) \Rightarrow \psi$ which stand for all ground instances obtained by substituting $x \in \texttt{Agents}$.
In these cases, the filtering is applied to the ground instances, and the same rule can exist for an agent and be filtered out for another.

\subsection{Contractualist Argumentation Theory}

We have described the process for incorporating the moral reasons used to construct contractualist arguments. This method for generating moral reasons can be applied to different argumentation systems, each with its own language $\mathcal{L}$ and rule set $\mathcal{R}$. However, we now propose a specific structure for creating Argumentation Theories that are compatible with contractualist reasoning. In this subsection, we define a \emph{Contractualist Argumentation Theory} (CAT) by introducing additional constraints on $\mathcal{K}$, $\mathcal{L}$ and $\mathcal{R}$. The central idea behind a CAT is to construct multiple arguments, each representing a moral reason of a single individual and concluding either in favour of performing or not performing an action. These arguments will then be compared to determine whether any one of them is sufficiently strong to defeat the opposing arguments.

To achieve this, a CAT must contain a literal $a$ such that $a \in \mathcal{L}$ and $a \notin \mathcal{K}$, where $a$ represents the action under evaluation. The same conditions must hold for $\neg a$. Additionally, neither $a$ nor $\neg a$ may appear among the premises of any rule. Formally: $\{r \in R \mid a \in \texttt{body}(r)\} = \emptyset$.

The purpose of this restriction is to use $a$ and $\neg a$ exclusively to define rebuttal attacks between the arguments of different agents. Consequently, they may appear as conclusions of arguments, but they cannot be used as premises for deriving further conclusions. To see why this matters, consider the example from Section~\ref{sec:example}: argument $A_{11}$ concludes $\neg a$ on the basis of an agent $\alpha$'s preferences, while $A_{12}$ concludes $a$ on the basis of agent $\beta$'s preferences. Because $a$ and $\neg a$ are the conclusions of these arguments, they stand in rebuttal with one another, and the defeat relation determines which prevails. If $a$ or $\neg a$ were instead permitted to appear as premises in further rules, an agent could derive additional conclusions from the fact that an action is performed or not, mixing the outcome of the moral evaluation back into the moral reasoning itself and producing circular or unintended inferences. The restriction ensures that the argumentation terminates cleanly at the point of moral judgment: the only role of these literals is to bring the arguments of different agents into direct conflict, making explicit that they are evaluating the same action from opposing standpoints. They are therefore present in the shared argumentation theory, but play no role in the construction of any individual agent's sub-arguments; they arise only at the final inferential step of each contractualist argument, where the moral reason of a single individual is connected to a verdict on the action.

We now define the moral arguments that will be compared, which we call \emph{contractualist arguments} \texttt{CArgs}. Informally, such arguments must conclude either $a$ or $\neg a$ and must contain at least one moral reason.
We model this through two properties, where \texttt{Sub} returns all sub-arguments of an argument.

\begin{itemize}
    \item $\forall C \in \texttt{CArgs},\ \texttt{Conc}(C) \in  \{a, \neg a\}$ 
    \item $\forall C \in \texttt{CArgs}, \exists p_v(\alpha) \in K_p \mid p_v(\alpha) \in \texttt{Sub}(C)$
\end{itemize}

In general, each contractualist argument $C$ contains a single moral reason $p_v(\alpha)$ and is therefore associated with one agent and one value. In addition to $p_v(\alpha)$, $C$ may contain any number of other atoms as needed. Since these additional atoms do not represent moral reasons, we recommend defining them as axioms in order to separate the moral reasoning process from factual or non-moral reasoning. Although this is not a strict requirement, it provides a practical way of implementing contractualism in \aspic.

The role of the axioms is to provide each contractualist argument with a structure that makes its corresponding moral objection legitimate. For example, suppose an action harms some agent $x$. In that case, another agent $y$ cannot raise a legitimate moral objection to the action solely on the basis of the physical harm suffered by $x$. Therefore, even if $y$ assigns sufficient importance to the violation of wellbeing and $p_{\textit{wellbeing}}(y)$ is included as an ordinary premise, the structure of the argument should prevent it from concluding either $a$ or $\neg a$. Section~\ref{sec:example} provides concrete examples of arguments combining axioms and moral reasons.

It is worth clarifying the sense in which contractualist arguments are truly \emph{arguments} rather than just logical derivations. The moral content of a CAT does not reside in the axioms or strict rules, which indeed carry no moral import of their own, but exclusively in the ordinary premises $p_v(\alpha)$. These premises are defeasible: they can be challenged, for instance through undercutting attacks that target the rules connecting them to the moral conclusion, or through rebutting attacks when a competing moral reason concludes the opposite verdict on the action. The axioms serve only to establish the structural preconditions that make a moral objection legitimate without themselves contributing moral weight to the argument. For instance, a privacy argument does not arise solely from the disclosure of sensitive information, but it also requires the owner and receiver of that information to be different agents. This argument would then be composed of an ordinary premise that contains moral information (sensitive information disclosure) and two non-moral axioms that check its validity (information owner and information receiver).

\subsection{Resolving Conflicts in CATs}

One the CAT has been created with multiple $C$, we should define a method for assessing how to resolve the rebuttal attacks and reach a conclusion -- whenever possible. VBFs define a subjective AT for each agent, and compute the intersection of the grounded extensions of each subjective AT, identifying the set of arguments accepted by all agents. However, for moral reasoning, such a method is often too conservative. Consider, for example, a situation in which an autonomous vehicle must choose between running over a pedestrian or turning and scratching the car. The pedestrian would have a very strong argument in favour of turning the car, whereas the car owner would only have a comparatively weak argument against it. Intuitively, the morally preferable action is clear, yet the intersection of the grounded extensions of the agents’ subjective ATs might fail to support either action. For this reason, we argue that an approach based on competing reasons is both compatible with contractualist reasoning and better suited for moral evaluation. 

As described above, our framework includes a single set $\mathcal{K}^*_p$ containing the moral reasons of all agents. Although these premises are subjective in nature, they are incorporated into a single ``objective'' AT. This allows the strengths of different arguments to be compared using the weights defined by $\texttt{ValProp}$. As we recommend to include only one $p_v(\alpha)$ in each argument, and each $p_v(\alpha)$ is associated with one weight $\texttt{ValProp}(\alpha,v,p)$, we can assign the same weight to the argument in the AT. We can then exploit the preference-based mechanisms for resolving attacks in \aspic. More specifically, we propose to assign the weight given by $\texttt{ValProp}(\alpha,v,p)$ to the top rule -- the last inference rule used -- of the corresponding argument. Because $\texttt{ValProp}$ is a totally ordered set, this naturally induces a preference pre-order over rules that follows the same ordering relation. We will make the indifference weight ``?'' the same order as the lowest weight. Finally, the comparison between arguments should be used using a last-link elitist principle, such that only the last rule is compared to compute defeats. Subarguments consist entirely of axioms and factual conclusions carrying no moral weight, so comparing arguments on the last-link alone is both sufficient and consistent with the contractualist prohibition on aggregating reasons. Extensions permitting multiple moral reasons within a single argument would require a more sophisticated treatment, which we leave to future work.

\section{Example}\label{sec:example}

In this section we will illustrate how our approach can model contractualist reasoning.
We consider a simple scenario that could take place in a household, and we will focus on two values: \textit{privacy} and \textit{autonomy}.

We suppose that an agent knows that a person $\alpha$ has secretly smoked a cigarette.
Another person $\beta$, perhaps the partner of the first, is now interacting with the agent, which finds itself having to decide whether to disclose the information or not.
We will call $a$ the action of disclosing, and consequently $\neg a$ the decision not to disclose.

Following the contractualist principles, we determine that this action affects two people, namely $\alpha$ and $\beta$.
We can see how the two values play a different role for each agent: $\alpha$ cares about privacy and would not want the information revealed; the action is not influencing $\alpha$'s autonomy, since this is not about keeping $\alpha$ from smoking.
On the other hand, there are no privacy concerns for $\beta$, while withholding the information would have a negative impact on $\beta$'s autonomy.
Note that these are not the only values at play, but we chose to focus only on these two for this example.

For simplicity, the function \texttt{ValLimit} is defined to be equal to $2$ for all agents and values.
For instance, \texttt{ValLimit}$(\alpha,\textit{privacy})=2$.

To represent whether $\alpha$ finds the information about the sneaky cigarette to be sensitive we define the ordinary premise $s$.
We recall that through \texttt{ValProp} function, each agent can assign a weight to propositions-values combinations; we model this in Table~\ref{tab:valprop}.

\begin{table}[h!]
    \centering
    \begin{tabular}{rcc}
         $s$               & $\alpha$ & $\beta$ \\
         \toprule
         \textit{privacy}  & 3      & 1         \\
         \textit{autonomy} & 2      & 2         \\
    \end{tabular}
    \caption{A representation of \texttt{ValProp} for the premise $s$, the values of \textit{privacy} and \textit{autonomy}, and the agents $\alpha$ and $\beta$.}
    \label{tab:valprop}
\end{table}

To model this problem, we will focus on once concern at the time, starting with information exchange.
The first aspect we consider is who is the owner of the information exchanged, and we say $owner(x)$ if $x$, one of the agents, is the owner.
We model in a similar fashion the concept of information receiver $receiver(x)$ and whether an agent provided explicit consent to the action, $consent(x)$.
Since in the context of this example we consider these as undisputable facts, we will add them to the set of axioms, indicating the correct agent.

\begin{multline}
    K_n = \{\neg consent(\alpha), owner(\alpha), \neg owner(\beta), \\
    \neg recevier(\alpha), receiver(\beta)\}
\end{multline}

To build the filtered $K^*_p$, we compare \texttt{ValProp} with \texttt{ValLimit} for each agent and value.
We write $s_\textit{privacy}(\alpha)$ the ordinary premise filtered according to $\alpha$'s preferences on \textit{privacy}.

\begin{equation}
    K^*_p = \{s_\textit{privacy}(\alpha), \neg s_\textit{privacy}(\beta), s_\textit{autonomy}(\alpha), s_\textit{autonomy}(\beta)\}
\end{equation}

We define three rules to model the concerns about \textit{privacy} in the context of information exchange.
The first rule generates a privacy concern $privConcern(x)$ raised by agent $x$ if it is the owner and not the receiver of the information.
The second forbids the action of disclosure if $x$ has a privacy concern and the considers the information sensitive.
The final rule adds an exception to the previous case, if $x$ has given explicit consent to the action.

\begin{equation}
\begin{split}
    r_1(x)&: owner(x), \neg receiver(x) \Rightarrow privConcern(x)      \\
    r_2(x)&: privConcern(x), s_\textit{privacy}(x) \Rightarrow \neg a   \\
    r_3(x)&: consent(x), owner(x) \Rightarrow \neg r_2(x)
\end{split}
\end{equation}

We apply the filtering to the rules; specifically, the only that can be filtered is $r_2(x)$, which we substitute with $r_2(\alpha)$ in the set of rules; $r_2(\beta)$ is filtered out.

\begin{equation}
    r_2(\alpha): privConcern(\alpha), s_{\emph{privacy}}(\alpha) \Rightarrow \neg a
\end{equation}

Finally, we can build the arguments as follows.

\begin{equation}
\begin{split}
    A_1&: \neg consent(\alpha)                                                  \\
    A_2&: owner(\alpha)                                                         \\
    A_3&: \neg owner(\beta)                                                     \\
    A_4&: \neg recevier(\alpha)                                                 \\
    A_5&: receiver(\beta)                                                       \\
    A_6&:s_\textit{privacy}(\alpha)                                             \\
    A_7&:\neg s_\textit{privacy}(\beta)                                         \\
    A_8&:s_\textit{autonomy}(\alpha)                                            \\
    A_9&:s_\textit{autonomy}(\beta)                                             \\
    A_{10}&: A_2, A_4 \overset{r_1(\alpha)}\Longrightarrow privConcern(\alpha)  \\
    A_{11}&: A_{10}, A_6 \overset{r_2(\alpha)}\Longrightarrow \neg a
\end{split}
\end{equation}

The conclusion of not performing the disclosure is not final: we will now extend our example by representing the \textit{autonomy} concerns in the same context: information exchange.
We add a rule to represent that the receiver of the information might see their autonomy increased if the information is shared (by enabling him to make better-informed choices).

\begin{equation}
    r_4(x): receiver(x), s_\textit{autonomy}(x) \Rightarrow a
\end{equation}

We apply the filter based on the accepted ordinary premises, obtaining both $r_4(\alpha)$ and $r_4(\beta)$.
We observe that we can now produce additional arguments.

\begin{equation}
    A_{12}: A_5, A_9 \overset{r_4(\beta)}\Longrightarrow a
\end{equation}

If we wanted to consider additional contexts -- for instance, the health risks associated with smoking -- we would now add the necessary axioms and rules to represent the concerns associated with each of the values considered.

Observing the generated arguments, we see that $A_{11}$ and $A_{12}$ are in conflict.
We apply the definition of ordering $\prec$, by looking at \texttt{ValProp}$(\alpha,\textit{privacy},s)$ and \texttt{ValProp}$(\beta,\textit{autonomy},s)$, reported in Table~\ref{tab:valprop}.
Since $3 \not < 2$, $A_{11}$ defeats $A_{12}$; we can now conclude $\neg a$ and thus not disclose the information.

\section{Discussion and Conclusion}\label{sec:discussion}

In this article, we proposed a contractualist framework for moral reasoning in \aspic{} that models moral evaluation as a hypothetical negotiation among affected agents, where each can raise reasons for or against it. These reasons are based on values, such as \textit{wellbeing}, and capture how strongly each individual perceives the action to promote or violate those values. The framework then builds arguments from these moral reasons while keeping factual reasoning separate from moral evaluation. To ensure compatibility with contractualism, each argument represents the perspective of a single individual and is grounded in principles that anyone in a similar situation could appeal to. Finally, the framework compares the competing arguments according to the strength of the underlying moral reasons in order to determine which objections are strongest and whether an action can be morally justified.

Contractualism is a promising framework for developing ethically aligned AI. It has been proposed as a strong candidate for value alignment \cite{Levine2025} and for the design of social robots \cite{Chater2023}. Furthermore, research in moral psychology suggests that people tend to trust both individuals and social robots more when they rely on contractualist reasoning rather than utilitarian or deontological approaches \cite{Everett2016,Gil-Buitrago2025}. Despite this potential, there has so far been only one attempt to develop a theoretical model for contractualism in AI \cite{Vidal2024} and another attempt to actually implement it \cite{Dalmasso2024}. The approach presented in the current paper offers a more flexible implementation by relying on values rather than norms. In addition, it is more faithful to Scanlon’s contractualism through its use of structured arguments defined in ASPIC+. Finally, because our framework focuses exclusively on moral reasoning, it can be combined with any method for extracting non-moral information from the context.

The approach presented in this paper shares a common motivation with VBFs~\cite{wallner2024value}: both aim to account for the role of values in shaping the arguments an agent is prepared to make, rather than using values solely to resolve conflicts between already-constructed arguments.
This stands in contrast to other approaches, such as value-based argumentation~\cite{atkinson2021value}, where values are associated to abstract arguments and ordered according to an audience's preferences, conditioning the effectiveness of attacks rather than the construction of arguments.
Our approach resonates naturally with Scanlon's contractualism: what matters is not only which reasons prevail, but which reasons each party is in a position to advance in the first place. 
An agent cannot reasonably reject a principle on grounds that fall outside their value profile, and the filtering mechanism gives this idea a formal counterpart.

However, rather than constructing separate per-agent theories, we allow the reasons of different agents to interact directly within a common framework, with the defeat relation -- informed by each agent's preferences on weights -- determining which reasons prevail.
This mirrors the contractualist procedure of virtual bargaining, where agents do not reason in isolation but confront one another's reasons and assess whether any could be reasonably rejected.

The framework proposed by Wyner and Zurek~\cite{wyner2024satisfaction} shares a common motivation with our approach in that both treat values as structuring the space of reasons agents can advance, rather than resolving conflicts between pre-built arguments. Their model filters propositions through agents' value profiles to construct a \texttt{PropBaseClean}, extended into a lattice of intermediate positions to model satisficing agreement; our use of \texttt{ValProp} and \texttt{ValLimit} is directly inspired by this. However, the two frameworks diverge significantly in aim and structure. Wyner and Zurek are concerned with reaching propositional agreement, allowing agents to sacrifice lower-weighted values to converge on a shared set of propositions. Our framework, by contrast, seeks to determine whether an action can be justified to all parties on the basis of reasons none could reasonably reject. This difference in purpose has a formal counterpart: in their model, values remain implicit private filters, never appearing as explicit objects in the argumentation process, since individuals rarely state their values directly in discourse. Our framework departs from this position by making values explicit as components of moral reasons of the form $p_v(\alpha)$, with weights assigned via \texttt{ValProp} directly governing the defeat relation. This is a formal necessity for contractualist reasoning, which requires that the reasons of different individuals -- grounded in specific values -- be placed in direct comparison. In practical deployments, however, value weights would typically be assigned by system designers or inferred from context rather than stated by agents themselves, preserving something of the indirection that Wyner and Zurek consider essential.

Several limitations of the current framework point to directions for future work.
For instance, the framework produces a one-shot verdict and does not model the iterative exchange of reasons that characterises virtual bargaining.
Extending the approach to a dialogue setting, where agents can challenge, rebut, and revise their positions over multiple turns, would bring the formal model closer to the contractualist ideal and to realistic human-robot interaction.
Evaluating a single action at a time does not account for sequences of actions or their cumulative moral impact.
Many domestic scenarios involve chains of decisions whose moral weight depends on prior choices, and extending the framework to handle sequential reasoning is a natural next step.

In \aspic{}, the winning argument is determined by the grounded extension of the shared argumentation framework, with the defeat relation induced by the ordering over arguments.
The connection between this ordering and the contractualist notion of reasonable rejection remains informal, and a more principled account of how argument strength maps onto the threshold of reasonable rejection is needed.
Furthermore, the current definition of this ordering relies directly on the preferences set by the agents; as a consequence, the agents could manipulate their preferences to when an \aspic{} attack becomes a defeat, and thus control which arguments can win.

Finally, reasoning in \aspic{} is known to be computationally hard in the general case.
In the future, we propose model the logic from scenario descriptions, through an analysis of which aspects play a role and which values influence each.
In practice, this means instantiating only the necessary arguments, for instance through a large language model, rather than enumerating all possible arguments; however, a systematic complexity analysis of the resulting framework remains to be carried out.
The framework does not discover new morally relevant values from the scenario itself, and its coverage is therefore bounded by the value ontology provided to it.
In such an implementation, it becomes fundamental to cover a large enough set of concerns, such as information exchange, health risks, psychological risks, et cetera.
Integrating richer, empirically grounded value taxonomies is a promising direction.

\section*{Acknowledgments}

Funded by the Horizon Europe VALAWAI project (grant agreement number 101070930).

\bibliographystyle{abbrv}
\bibliography{references}

@incollection{wallner2024value,
  title={Value-based Reasoning in ASPIC+},
  author={Wallner, Johannes P and Wyner, Adam and Zurek, Tomasz},
  booktitle={Computational Models of Argument: Proceedings of COMMA 2024},
  pages={325--336},
  year={2024},
  publisher={SAGE Publications 1 Oliver's Yard, 55 City Road, London, EC1Y 1SP}
}

@article{modgil2014aspic,
  title={The ASPIC+ framework for structured argumentation: a tutorial},
  author={Modgil, Sanjay and Prakken, Henry},
  journal={Argument \& Computation},
  volume={5},
  number={1},
  pages={31--62},
  year={2014},
  publisher={SAGE Publications Sage UK: London, England}
}

@book{Scanlon1998,
   author = {Thomas Scanlon},
   city = {Cambridge},
   publisher = {Belknap Press of Harvard University Press},
   title = {What We Owe to Each Other},
   year = {1998}
}

@book{Scanlon2008,
   author = {Thomas Scanlon},
   city = {Cambridge, Mass.},
   publisher = {Belknap Press of Harvard University Press},
   title = {Moral Dimensions: Permissibility, Meaning, Blame},
   year = {2008}
}

@book{RAWLS1971,
   author = {John Rawls},
   doi = {10.2307/j.ctvjf9z6v},
   isbn = {9780674042605},
   month = {3},
   publisher = {Harvard University Press},
   title = {A Theory of Justice},
   year = {1971}
}

@article{Mackie1977,
   author = {J. L. Mackie},
   doi = {10.2307/2184791},
   issn = {00318108},
   issue = {1},
   journal = {The Philosophical Review},
   month = {1},
   pages = {134},
   publisher = {JSTOR},
   title = {Ethics: Inventing Right and Wrong},
   volume = {88},
   year = {1977}
}

@article{atkinson2021value,
  title={Value-based argumentation},
  author={Atkinson, Katie and Bench-Capon, Trevor},
  journal={Journal of Applied Logics},
  volume={8},
  number={6},
  pages={1543--1588},
  year={2021}
}

@article{Levine2024,
   author = {Sydney Levine and Nick Chater and Joshua B. Tenenbaum and Fiery Cushman},
   doi = {10.1017/S0140525X24001067},
   issn = {0140-525X},
   journal = {Behavioral and Brain Sciences},
   pages = {1-38},
   publisher = {Cambridge University Press},
   title = {Resource-rational contractualism: A triple theory of moral cognition},
   year = {2024}
}

@article{Misyak2014,
   author = {Jennifer B. Misyak and Nick Chater},
   doi = {10.1098/rstb.2013.0487},
   issn = {14712970},
   issue = {1655},
   journal = {Philosophical Transactions of the Royal Society B: Biological Sciences},
   month = {11},
   pmid = {25267828},
   publisher = {The Royal Society},
   title = {Virtual bargaining: a theory of social decision-making},
   volume = {369},
   year = {2014}
}

@article{Chater2022,
   author = {Nick Chater and Hossam Zeitoun and Tigran Melkonyan},
   doi = {10.1037/rev0000343},
   journal = {Psychological Review},
   title = {The Paradox of Social Interaction: Shared Intentionality, We-Reasoning, and Virtual Bargaining},
   year = {2022}
}

@article{Misyak2014b,
   author = {Jennifer B. Misyak and Tigran Melkonyan and Hossam Zeitoun and Nick Chater},
   doi = {10.1016/j.tics.2014.05.010},
   issn = {1364-6613},
   issue = {10},
   journal = {Trends in Cognitive Sciences},
   month = {10},
   pages = {512-519},
   pmid = {25073460},
   publisher = {Elsevier Current Trends},
   title = {Unwritten rules: virtual bargaining underpins social interaction, culture, and society},
   volume = {18},
   year = {2014}
}

@article{abbo2025concernsvalues,
  author={Abbo, Giulio Antonio
  and Belpaeme, Tony
  and Spitale, Micol},
  title={Concerns and Values in Human-Robot Interactions: A Focus on Social Robotics},
  journal={International Journal of Social Robotics},
  year={2026},
  month={Jan},
  day={14},
  volume={18},
  number={1},
  pages={4},
  issn={1875-4805},
  doi={10.1007/s12369-025-01351-1},
  url={https://doi.org/10.1007/s12369-025-01351-1}
}

@inproceedings{abbo2026multimodal-large-language,
author = {Abbo, Giulio Antonio and Lenaerts, Senne and Belpaeme, Tony},
title = {Multimodal Large Language Models for Real-Time Situated Reasoning},
year = {2026},
isbn = {9798400723216},
publisher = {Association for Computing Machinery},
address = {New York, NY, USA},
url = {https://doi.org/10.1145/3776734.3796242},
doi = {10.1145/3776734.3796242},
booktitle = {Companion Proceedings of the 21st ACM/IEEE International Conference on Human-Robot Interaction},
pages = {1161–1163},
numpages = {3},
location = {Edinburgh, Scotland, UK},
series = {HRI Companion '26}
}

@article{Vidal2024,
   author = {Luis Marcos-Vidal and Serena Marchesi and Agnieszka Wykowska and Clara Pretus},
   doi = {10.31234/osf.io/52x74},
   journal = {PsyArXiv},
   month = {7},
   publisher = {OSF},
   title = {Moral agents as relational systems: The Contract-Based Model of Moral Cognition for AI},
   year = {2024}
}

@inproceedings{Gil-Buitrago2025,
   author = {Helena Gil-Buitrago and Luis Marcos-Vidal and Clara Pretus},
   doi = {10.31234/osf.io/xjmve\_v1},
   title = {Does Contractualism Shape Trust and Perceived Agency in Social Robots?},
   editor="Osman, Nardine and Steels, Luc",
    booktitle="Value Engineering in Artificial Intelligence",
    year="2026",
    publisher="Springer Nature Switzerland",
    address="Cham",
    isbn="978-3-032-31055-2"
}

@article{Everett2016,
   author = {Jim A.C. Everett and David A. Pizarro and M. J. Crockett},
   doi = {10.1037/xge0000165},
   issn = {00963445},
   issue = {6},
   journal = {Journal of Experimental Psychology: General},
   month = {6},
   pages = {772-787},
   pmid = {27054685},
   publisher = {American Psychological Association Inc.},
   title = {Inference of Trustworthiness From Intuitive Moral Judgments},
   volume = {145},
   year = {2016}
}

@article{Chater2023,
    author = {Chater, Nick},
    title = {How could we make a social robot? A virtual bargaining approach},
    journal = {Philosophical Transactions of the Royal Society A: Mathematical, Physical and Engineering Sciences},
    volume = {381},
    number = {2251},
    pages = {20220040},
    year = {2023},
    month = {06},
    issn = {1364-503X},
    doi = {10.1098/rsta.2022.0040}
}

@article{Levine2025,
   author = {Sydney Levine and Matija Franklin and Tan Zhi-Xuan and Secil Yanik Guyot and Lionel Wong and Daniel Kilov and Yejin Choi and Joshua B. Tenenbaum and Noah Goodman and Seth Lazar and Iason Gabriel},
   month = {6},
   title = {Resource Rational Contractualism Should Guide AI Alignment},
   doi = {10.48550/arXiv.2506.17434},
   journal = {International Association for Safe \& Ethical AI (IASEAI)},
   year = {2026}
}

@inproceedings{Dalmasso2024,
author = {Dalmasso, Giovanni and Marcos-Vidal, Luis and Pretus, Clara},
title = {Modelling Moral Decision-Making in a Contractualist Artificial Agent},
year = {2024},
isbn = {978-3-031-85462-0},
publisher = {Springer-Verlag},
address = {Berlin, Heidelberg},
url = {https://doi.org/10.1007/978-3-031-85463-7_10},
doi = {10.1007/978-3-031-85463-7_10},
booktitle = {Value Engineering in Artificial Intelligence: Second International Workshop, VALE 2024, Santiago de Compostela, Spain, October 19–24, 2024, Revised Selected Papers},
pages = {155–175},
numpages = {21},
location = {Santiago de Compostela, Spain}
}

@article{dung1995acceptability,
  title={On the acceptability of arguments and its fundamental role in nonmonotonic reasoning, logic programming and n-person games},
  author={Dung, Phan Minh},
  journal={Artificial intelligence},
  volume={77},
  number={2},
  pages={321--357},
  year={1995},
  publisher={Elsevier}
}

@inproceedings{zurek2022towards,
  author       = {Adam Wyner and
                  Tomasz Zurek},
  editor       = {Nardine Osman and
                  Luc Steels},
  title        = {Towards a Formalisation of Motivated Reasoning and the Roots of Conflict},
  booktitle    = {Value Engineering in Artificial Intelligence - First International
                  Workshop, {VALE} 2023, Krakow, Poland, September 30, 2023, Proceedings},
  series       = {Lecture Notes in Computer Science},
  pages        = {28--45},
  publisher    = {Springer},
  year         = {2023},
  url          = {https://doi.org/10.1007/978-3-031-58202-8\_3},
  doi          = {10.1007/978-3-031-58202-8\_3},
  bibsource    = {dblp computer science bibliography, https://dblp.org}
}

@inproceedings{wyner2024satisfaction,
  author       = {Adam Z. Wyner and
                  Tomasz Zurek},
  editor       = {Nardine Osman and
                  Luc Steels},
  title        = {Satisfaction in Negotiation by Structured Values and Propositions},
  booktitle    = {Value Engineering in Artificial Intelligence - Second International
                  Workshop, {VALE} 2024, Santiago de Compostela, Spain, October 19-24,
                  2024, Revised Selected Papers},
  series       = {Lecture Notes in Computer Science},
  pages        = {176--192},
  publisher    = {Springer},
  year         = {2024},
  url          = {https://doi.org/10.1007/978-3-031-85463-7\_11},
  doi          = {10.1007/978-3-031-85463-7\_11},
  bibsource    = {dblp computer science bibliography, https://dblp.org}
}
\balance

\end{document}